# Predicting accurate probabilities with a ranking loss

**Aditya Krishna Menon**[1]                                                                                                            AKMENON@UCSD.EDU
**Xiaoqian Jiang**[1]    XLJIANG@UCSD.EDU
**Shankar Vembu**[2]    SHANKAR.VEMBU@UTORONTO.CA
**Charles Elkan**[1]    ELKAN@UCSD.EDU
**Lucila Ohno-Machado**[1]    MACHADO@UCSD.EDU

[1]University of California, San Diego, 9500 Gilman Drive, La Jolla CA 92093, USA
[2]University of Toronto, 160 College Street, Toronto, ON M5S 3E1, Canada

## Abstract

In many real-world applications of machine learning classifiers, it is essential to predict the probability of an example belonging to a particular class. This paper proposes a simple technique for predicting probabilities based on optimizing a *ranking loss*, followed by isotonic regression. This semi-parametric technique offers both good ranking and regression performance, and models a richer set of probability distributions than statistical workhorses such as logistic regression. We provide experimental results that show the effectiveness of this technique on real-world applications of probability prediction.

## 1. Introduction

Classification is the problem of learning a mapping from examples to labels, with the goal of categorizing future examples into one of several classes. However, many real-world applications instead require that we estimate the *probability* of an example having a particular label. For example, when studying the click behaviour of ads in computational advertising, it is essential to model the probability of an ad being clicked, rather than just predicting whether or not it will be clicked (Richardson et al., 2007). Accurate probabilities are also essential for medical screening tools to trigger early assessment and admission to an ICU (Subbe et al., 2001).

In this paper, we propose a simple semi-parametric model for predicting accurate probabilities that uses *isotonic regression* in conjunction with scores derived from optimizing a *ranking loss*. We analyze theoretically and empiri-cally where our approach can provide more reliable estimates than standard statistical workhorses for probability estimation, such as logistic regression. The model attempts to achieve good ranking (in an area under ROC sense) and regression (in a squared error sense) performance simultaneously, which is important in many real-world applications (Sculley, 2010). Further, our model is much less expensive to train than full-blown nonparametric methods, such as kernel logistic regression. It is thus an appealing choice in situations where parameteric models are employed for probability estimation, such as medical informatics and credit scoring.

The paper is organized as follows. First, we provide motivating examples for predicting probabilities, and define the fundamental concept of proper losses. We then review existing methods used to predict probabilities, and discuss their limitations. Next, we detail our method to estimate probabilities, based on optimizing a ranking loss and feeding the results into isotonic regression. Finally, we provide experimental results on real-world datasets to validate our analysis and to test the efficacy of our method.

We first fix our notation. We focus on probability estimation for examples $x \in \mathcal{X}$ with labels $y \in \{0, 1\}$. Each $x$ has a conditional probability function $\eta(x) := \Pr[y = 1|x]$. For our purposes, a model is some deterministic mapping $\hat{s}: \mathcal{X} \to \mathbb{R}$. A probabilistic model $\hat{\eta}$ is a model whose outputs are in $[0, 1]$, and may be derived by composing a model with a link function $f : \mathbb{R} \to [0, 1]$. The scores of a model may be thresholded to give a classifier $\hat{y} : \mathcal{X} \to \{0, 1\}$. We assume $\hat{s}$ is learned from a training set $\{(x_i, y_i)\}_{i=1}^n$ of $n$ iid draws from $\mathcal{X} \times \{0, 1\}$.

## 2. Background and motivation

Classically, the supervised learning literature has focussed on the scenario where we want to minimize the number of misclassified examples on test data. However, practical applications of machine learning models often have more complex constraints and requirements, which demand that we output the probability of an example possessing a label. Examples of such applications include:

**Building meta-classifiers**, where the output of a model is fed to a meta-classifier that uses additional domain knowledge to make a prediction. For example, doctors prefer to use a classifier's prediction as evidence to aid their own decision-making process (Manickam & Abidi, 1999). In such scenarios, it is essential that the classifier assess the *confidence* in its predictions being correct, which may be captured using probabilities;

**Using predictions to take actions**, such as deciding whether or not to contact a person for a marketing campaign. Such actions have an associated utility that is to be maximized, and maximization of expected utility is most naturally handled by estimating probabilities rather than making hard decisions (Zadrozny & Elkan, 2001);

**Non-standard learning tasks**, where problem constraints demand estimating uncertainty. For example, in the task of learning from only positive and unlabelled examples, training a probabilistic model that distinguishes labelled versus unlabelled examples is a provably (under some assumptions) sufficient strategy (Elkan & Noto, 2008).

Intuitively, probability estimates $\hat{\eta}(\cdot)$ are accurate if, on average, they are close to the true probability $\eta(\cdot)$. Quantifying "close to" requires picking some sensible discrepancy measure, and this idea is formalized by the theory of proper loss functions, which we now discuss. A model for binary classification uses a loss function $\ell : \{0,1\} \times \mathbb{R} \to \mathbb{R}_+$ to measure the discrepancy between a label $y$ and the model's prediction $\hat{s}$ for some example $x$. If our model outputs probability estimates $\hat{\eta}$ by transforming scores with a link function $f(\cdot)$, we may equivalently think of there being a probabilistic loss $\ell^P(\cdot,\cdot)$ such that $\ell(y,\hat{s}) = \ell^P(y, f(\hat{s}))$. The empirical error of $\hat{s}$ with respect to the loss $\ell$ is

$$\mathcal{E}_{\text{emp}}(\hat{s}(\cdot)) = \frac{1}{n} \sum_{i=1}^{n} \ell(y_i, \hat{s}(x_i)),$$

which is a surrogate for the generalization error

$$\begin{aligned}
\mathcal{E}(\hat{s}(\cdot)) &= \mathbb{E}_x \mathbb{E}_{y|x} \ell(y, \hat{s}(x)) \\
&= \mathbb{E}_x [\eta(x)\ell(1, \hat{s}(x)) + (1-\eta(x))\ell(0, \hat{s}(x))] \\
&:= \mathbb{E}_x L_\ell(\eta(x), \hat{s}(x)).
\end{aligned} \quad (1)$$

The term $L_\ell(\eta, \hat{s})$ is a measure of discrepancy between an example's probability of being positive and its predicted score. Let $s^*(\eta) = \operatorname{argmin}_s L_\ell(\eta, s)$. Then, we call a loss function $\ell$ *Bayes consistent* (Buja et al., 2005) if for every $\eta \in [0,1]$, $s^*(\eta) \cdot (\eta - \frac{1}{2}) \geq 0$, meaning that we have the same sign as the optimal prediction under the 0-1 loss $\ell(y,\hat{s}) = \mathbf{1}[y\hat{s} \leq 0]$. If $s^*(\eta)$ is invertible, then $(s^*)^{-1}(s^*(\eta)) = \eta$, so that the optimal scores are some transformation of $\eta(x)$. In such cases, we call the corresponding probabilistic loss $\ell^P$ a *proper* (or *Fisher-consistent*) loss (Buja et al., 2005), and say that $\ell$ corresponds to a proper loss.

Many commonly used loss functions, such as square loss $\ell(y,\hat{s}) = (y - \hat{s})^2$, and logistic loss $\ell(y,\hat{s}) = \log(1 + e^{-(2y-1)\hat{s}})$, correspond to a proper loss function. Thus, a model with good regression performance according to squared error, say, can be thought to yield meaningful probability estimates. The hinge loss of SVMs, $\ell(y,\hat{s}) = \max(0, 1 - (2y-1)\hat{s})$, is Bayes consistent but does *not* correspond to a proper loss function, which is why SVMs do not output meaningful probabilities (Platt, 1999).

## 3. Analysis of existing paradigms to learn accurate probabilities

We now analyze two major paradigms for probability estimation, and study their possible failure modes.

### 3.1. Optimization of a proper loss

A direct approach to predicting probabilities is to optimize a proper loss function on the training data using some hypothesis class, e.g. linear separators. Examples include logistic regression and linear regression (after truncation to $[0,1]$), which are instances of the generalized linear model framework, which assumes $\mathbb{E}[y|x] = f(w^T x)$ for some link function $f(\cdot)$. The loss-dependent error measure, $L_\ell(\eta, \hat{s})$, is one metric by which we can choose amongst proper losses. For example, the discrepancy measures for square and logistic loss are (Zhang, 2004)

$$L_{\text{square}}(\eta, \hat{s}) = (\eta - \hat{s})^2 + C_1 \quad (2)$$

$$L_{\text{logistic}}(\eta, \hat{s}) = \text{KL}\left(\eta \Big|\Big| \frac{1}{1+e^{-\hat{s}}}\right) + C_2, \quad (3)$$

where KL denotes the Kullback-Leibler divergence, and $C_1, C_2$ are independent of the prediction $\hat{s}$. Based on this, Zhang (2004) notes that logistic regression has difficulty when $\eta(x)(1 - \eta(x)) \approx 0$ for some $x$, by virtue of requiring $|\hat{s}(x)| \to \infty$. This has been observed in practical uses of logistic regression with imbalanced classes (King & Zeng, 2001; Foster & Stine, 2004), with the latter proposing the use of linear regression as a more robust alternative.

### 3.2. Post-processing methods

A distinct strategy is to train a model in some manner, and then extract probability estimates from it in a post-processing step. Three popular techniques of this type are Platt scaling (Platt, 1999), binning (Zadrozny & Elkan, 2001), and isotonic regression (Zadrozny & Elkan, 2002). We focus on the latter, as it is more flexible than the former two approaches by virtue of being nonparametric, and has been shown to work well empirically for a range of input models (Niculescu-Mizil & Caruana, 2005).

Isotonic regression is a nonparametric technique to find a monotone fit to a set of target values. In a learning context, the method was used in (Zadrozny & Elkan, 2002) to learn meaningful probabilities from the scores of an input model. Mathematically, suppose we have predictions $\{\hat{s}_i\}_{i=1}^n$ from some input model, with corresponding true labels $\{y_i\}_{i=1}^n$, and WLOG suppose that $\hat{s}_1 \leq \hat{s}_2 \leq \ldots \leq \hat{s}_n$. Then, isotonic regression learns scores $\{\tilde{s}_i\}_{i=1}^n$ via the optimization

$$\min_{\tilde{s}_1,\ldots,\tilde{s}_n} \sum_{i=1}^n (y_i - \tilde{s}_i)^2 : \tilde{s}_i \leq \tilde{s}_{i+1} \quad \forall i \in \{1,\ldots,n-1\}.$$

This finds the best monotone fit to the training labels (as ordered by the input model's scores) in a squared loss sense. (In fact, the optimal solution will minimize any proper loss (Brümmer & Preez, 2007).) If the input scores $\{\hat{s}_i\}$ are sorted, then there is an $O(n)$ algorithm to solve this problem, called *pool adjacent violators* (PAV) (Barlow et al., 1972).

When $y_i \in \{0,1\}$, it is easy to verify that $\tilde{s}_i \in [0,1]$, so that the result is a probabilistic model. Indeed, isotonic regression can be thought of as nonparametrically learning a monotone link function $f(\cdot)$ to create a probabilistic model $f(\hat{s}(\cdot))$. However, the resulting model is only defined on the training examples, and we need to define some interpolation scheme to make predictions on future examples. One natural scheme is a linear interpolation between the training scores (Cosslett, 1983). Observe that isotonic regression preserves the ordering of the input model's scores, although potentially introducing ties i.e. $f(\hat{s}(\cdot))$ is not injective. To break ties on training examples, we may simply refer to the corresponding original model's scores. Linear interpolation breaks most ties[1] on test examples.

### 3.3. Possible failure modes

There are at least two main reasons why the above paradigms may not yield accurate probabilities:

**Misspecification**. In practice, simple models based on parametric assumptions will often be misspecified: for example, logistic regression assumes the parametric form $\eta(x) = 1/(1+e^{-w^T x})$ for some $w$, but this assumption may not always hold. While we cannot learn $\eta(x)$ if we cannot represent it in our hypothesis class, Equation 1 says that our model's predictions will in expectation be close to $\eta(x)$ according to some discrepancy measure. It is possible for a model like logistic regression to be correctly specified up to the choice of link function, i.e. $\eta(x) = f(w^T x)$, but $f(\cdot)$ is not the sigmoid function. The maximum likelihood estimates of a generalized linear model with a misspecified link function are known to be asymptotically biased (Czado & Santner, 1992). Isotonic regression alleviates this particular type of misspecification, but is still vulnerable if its input scores are misspecified.

A natural defense against misspecification is using a nonparametric method such as kernel logistic regression (KLR). This model will be able to learn any measurable $\eta(x)$ with a universal kernel (Zhang, 2004). In many practical applications, such methods are seen as too expensive to both train (requiring $O(n^3)$ time (Zhu & Hastie, 2005)) and test (requiring $O(n)$ time to make a prediction, since the weights on training examples generally have full support, unlike a kernel SVM).

**Finite-sample effects**. When optimizing an unregularized proper loss on a finite training set of $n$ examples, the probability estimates may be biased. Indeed, the finite sample MLE for the parameters of a generalized linear model (such as logistic regression) have a bias of $O(1/n)$ (Cordeiro & McCullagh, 1991), and thus the probability estimates are also biased. King and Zeng (2001) show that the constant in the $O(\cdot)$ depends on the *imbalance* in the classes, meaning that logistic regression can give biased probability estimates when attempting to model a rare event. It is possible to perform bias correction explicitly via a post-hoc modification of the learned parameters (King & Zeng, 2001), or implicitly by choosing a Jeffrey's prior regularizer (Firth, 1993).

Similarly, isotonic regression may overfit even if the input scores give a good ranking on test data. This can happen when there are "gaps" amongst the input scores. The simplest example is when the largest input score $\hat{s}_{\max}$ is associated with a positive label. Assuming there is only one example with this score, isotonic regression will predict the probability for *any* test example with score $\geq \hat{s}_{\max}$ to be 1, which is too optimistic and will likely be a poor model in this region of input space. The problem arises because we have insufficient representation of scores in $[\hat{s}_{\max}, \infty)$.

## 4. Extracting probabilities from a ranker

The semi-parametric route of isotonic regression is appealing because it involves a simple post-processing step, while strictly enhancing the hypothesis class of the input model. For this reason, we focus on this semi-parametric paradigm in what follows. Our hope is to design a model that is at least as accurate, and not much more difficult to train than workhorses such as logistic regression.

To use isotonic regression to get accurate estimates, we must specify *what* scores we will feed it as input. We may thus ask what characteristics such scores should possess so as to yield accurate probability estimates. We make the simple observation that isotonic regression interacts with the scores of the input model in only one way: it uses them to enforce the monotonicity constraint on the output. Thus, intuitively, isotonic regression will perform well when the (pairwise) *ranking* of the original scores is good, and so this should be our objective when training our input model. We now attempt to formalize this intuition, and present our proposed method.

---

[1] If the training example with largest score has corresponding isotonic regression prediction of 1, every test example with a larger score will also have a prediction of 1.

### 4.1. Isotonic regression and ranking performance

The real-valued score that a model assigns to each example may be used to rank examples according to confidence of having a positive label. The pairwise ranking performance of a model may be measured using the *area under the ROC curve* (AUC), being the probability that a randomly drawn positive example has a higher score than a randomly drawn negative example. It is formally defined below.

**Definition 1.** *(Clémençon et al., 2006) The AUC $\mathscr{A}(\hat{s}(\cdot))$ of a model $\hat{s}: \mathscr{X} \to \mathbb{R}$ is*

$$\mathscr{A}(\hat{s}(\cdot)) = \Pr_{(x_1,y_1),(x_2,y_2)}[\hat{s}(x_1) \geq \hat{s}(x_2)|y_1 = 1, y_2 = 0].$$

We henceforth think of a model $\hat{s}(\cdot)$ as equivalently representing a ranker of examples. A natural quantity to study is the model $\hat{s}(\cdot)$ that induces the Bayes-optimal ranker, meaning $\mathscr{A}(\hat{s}(\cdot)) \geq \mathscr{A}(\tilde{s}(\cdot))$. Intuitively, we expect this optimal ranker to be $\eta(x)$, or some (strictly) monotone transform $c(\cdot)$ thereof, and indeed this may be proven (Clémençon et al., 2006). Therefore, finding accurate probabilities can conceptually be cast as finding accurate ranking, and then recovering the correct transformation $c(\cdot)$.

We may now show that isotonic regression applied to a Bayes-optimal ranker (in the sense of AUC performance) will recover the true probabilities, by inferring the $c(\cdot)$ discussed above. This can be proven by observing that isotonic regression returns *calibrated* scores (see e.g. (Kalai & Sastry, 2009) for a proof). Calibration of probability estimates is defined as follows.

**Definition 2.** *(Schervish, 1989) We say that a model $\hat{s}$ is calibrated if, for every $\alpha \in \hat{s}[\mathscr{X}]$, $\alpha = \Pr[y = 1|\hat{s} = \alpha]$.*

We now show that calibration and Bayes-optimal AUC performance implies accuracy of estimates.

**Proposition 1.** *Let the model $\hat{s}$ be a Bayes-optimal ranker, meaning $\mathscr{A}(\hat{s}(\cdot)) \geq \mathscr{A}(\tilde{s}(\cdot))$ for every model $\tilde{s}$. Then, if $\hat{s}$ is calibrated, $\hat{s}(x) = \eta(x)$ for all $x$.*

*Proof.* Recall that for an optimal ranker, $\hat{s}(x) = c(\eta(x))$ for some strictly monotone $c(\cdot)$. Let $S = \hat{s}[\mathscr{X}]$. If $\hat{s}$ is calibrated, then by definition

$$\Pr[y = 1|c(\eta(x)) = s] = s, \quad \forall s \in S.$$

Any strictly monotone transformation $c(\cdot)$ must have an inverse $c^{-1}(\cdot)$. Thus the above may be rewritten as

$$\Pr[y = 1|\eta(x) = c^{-1}(s)] = s, \quad \forall s \in S.$$

But we know that $\eta(x)$ is a calibrated predictor:

$$\Pr[y = 1|\eta(x) = c^{-1}(s)] = c^{-1}(s), \quad \forall s \in S.$$

Therefore, $c^{-1}(s) = s$ for all $s$, meaning $c(s) = s$, and thus, $\hat{s}(x) = \eta(x)$. □

### 4.2. Our proposal: ranking loss + isotonic regression

The above suggests a natural idea: directly optimize the AUC on the training set, and post-process its scores with isotonic regression. This can be viewed as learning a model that has good ranking performance (by virtue of first optimizing a ranking loss) as well as good probability estimation performance (by virtue of isotonic regression optimizing every proper loss). With appropriate handling of ties, isotonic regression enforces strict monotonicity, and so its scores will have the same AUC as the original model. On a finite training set with $n^+$ positive and $n^-$ negative examples, the empirical AUC $\mathscr{A}_{\text{emp}}$ can be computed as

$$\mathscr{A}_{\text{emp}} = \frac{1}{n^+ n^-} \sum_{i,j} \mathbf{1}[\hat{s}(x_i) \geq \hat{s}(x_j)] y_i (1 - y_j), \quad (4)$$

which can be seen to measure the number of *concordant pairs* in the training set i.e. pairs of examples where the predicted scores respect the ordering according to the label.

To maximize AUC, we may follow the pairwise ranking framework (Herbrich et al., 2000; Joachims, 2002), which uses a regularized convex approximation to the RHS of Equation 4:

$$\min_w \sum_{i,j} \ell(\hat{s}(x_j;w) - \hat{s}(x_i;w), 1) y_i (1 - y_j) + \lambda \Omega(w), \quad (5)$$

where $\ell(\cdot, \cdot)$ is some convex loss function, and $\Omega(\cdot)$ is a regularization function with strength $\lambda > 0$. We use a linear scoring function[2] i.e. $\hat{s}(x;w) = w^T x$, for which the regularizer is generally taken to be the $\ell_2$ norm $\frac{1}{2}||w||_2^2$. While the above loss function nominally requires $O(n^2)$ time to compute the gradient, clever algorithms can speed this up (Joachims, 2006). Empirically, it has been observed that stochastic gradient descent on the objective converges in a fraction of an epoch (Sculley, 2009).

---

[2] The ranker may of course be kernelized, but in this case there is no clear reason to eschew kernel logistic regression.

The issue of how best to maximize AUC is not settled. For example, Kotlowski et al. (2011) show that the ranking error (viz. $1 - \mathscr{A}$) of a model can be upper bounded by its balanced logistic loss (viz. the logistic loss balanced by the respective class priors), suggesting that in practice one may approximately maximize AUC using logistic regression. (We say "approximately" because the result only provides a lower bound on the resulting AUC.) Consequently, post-processing the output of logistic regression with an isotonic regression fit is a worthwhile strategy to explore, and is indeed something we look at in our experiments. (Results such as (King & Zeng, 2001) suggest that logistic regression is not appropriate for imbalanced data because its raw *probabilities* are biased, not its *ranking* of examples.)

### 4.3. Justification of model

Our model operates by finding some $\hat{s}(x) = w^T x$ that optimizes Equation 5, and then post-processing these scores with isotonic regression. To argue that this model learns something meaningful, we need to show two things: (a) the solution to the convex optimization problem of Equation 5 will (asymptotically) yield a Bayes-optimal ranker, assuming the model is correctly specified, and (b) isotonic regression on top of a Bayes-optimal ranker will recover $\eta(x)$. Point (a) can be established if the underlying classification model uses a universal kernel (Clémençon et al., 2006). For a linear kernel, this means that we can learn the optimal ranking if the underlying probability is of the form $c(w^T x)$ for some monotone increasing $c(\cdot)$. Point (b) was established in Section 4.1, and it is further the case that the isotonic regression estimate on a finite training set is consistent, under mild regularity assumptions (Brunk, 1958).

If our model is misspecified – that is, $\eta(x)$ is not a monotone transformation of $w^T x$ – then the above analysis does not hold: the optimal ranker and the optimal regressor within our hypothesis class may be different. We can however show the following weaker result about the *empirical* squared error resulting from our isotonic regression step.

**Proposition 2.** *Suppose a model $\hat{s} \colon \mathscr{X} \to \mathbb{R}$ has empirical AUC $\mathscr{A}_{emp}$ on a training set with empirical base rate $\hat{\pi}$. Then, there is a model $\tilde{s}$ with the same empirical AUC, and empirical square loss at worst $\frac{1}{2}\sqrt{\hat{\pi}(1-\hat{\pi})(1-\mathscr{A}_{emp})}$.*

*Proof.* We previously established that isotonic regression will maintain the empirical AUC, and so we focus on the resulting squared error. Recall that the empirical AUC penalizes the number of discordant positive and negative example pairs. We may rewrite it as $\mathscr{A}_{\text{emp}} = 1 - k/n^+n^-$, so that there are $k$ discordant pairs. Suppose these pairs arise due to $a$ positive and $b$ negative examples, $a \leq n^+, b \leq n^-$. The worst placement of these pairs is if *all* the $a$ positives have lower scores than the $b$ negatives. In this case, we have $k = ab$, and it is easy to check that the resulting square loss is $\frac{1}{n}\frac{ab}{(a+b)}$. This score is largest when $a^* = \min(n^+, \lceil\sqrt{k}\rceil)$, where it attains the value $\frac{ka^*}{n(k+(a^*)^2)}$. This may be bounded by $\frac{\sqrt{k}}{2n}$, and so the worst possible square loss for isotonic regression is $\frac{\sqrt{n^+n^-}}{2(n^++n^-)}\sqrt{1-\mathscr{A}_{\text{emp}}}$, proving the claim. □

Since the empirical AUC is concentrated around the true AUC (Agarwal et al., 2005), the above is easily extended to a bound in terms of the true AUC. However, this is still a bound on the *training* squared error, and so is not a true generalization bound.

### 4.4. Comparison to existing methods

The first step of our method attempts to maximize the pairwise ranking performance, and the isotonic regression step attempts to achieve low squared error. By construction, then, our method attempts to achieve both good ranking and regression (in a squared error sense) performance. Good performance in both metrics is important in many applications, such as computational advertising (Richardson et al., 2007). The idea of learning models with good ranking and regression performance was proposed in the combined regression and ranking (CRR) framework of Sculley (2010). A similar model for logistic loss was proposed by Ertekin and Rudin (2011). The basic idea of such an approach is to simultaneously optimize the ranking and regression losses in a parametric manner, by minimizing a linear combination of both losses. The hope is that this yields "best of both worlds" performance in these objectives. Empirically, Sculley (2010) observed that generally the AUC obtained from such an approach was no worse than that of optimizing the ranking loss alone, while in some cases there was an improvement in the regression performance. By contrast, while we do make a parametric assumption for the ranking loss, our regression component is nonparametric and hence more powerful. Thus, in light of Sculley (2010)'s finding, we expect to achieve equitable ranking performance to methods like CRR, and better regression performance.

As the previous section makes clear, the idea of post-processing scores with isotonic regression is not new. However, to our knowledge, prior work has not studied the implications of applying this processing to a model that optimizes ranking performance; the idea is hinted at in (Sculley et al., 2011), but not discussed formally. Indeed, we argue that the scores from optimizing a ranking loss are the "correct" ones to use as input to isotonic regression, in the sense of recovering the true probability when the ranker is correctly specified. (Previous work has looked at applying isotonic regression to a general ranker that assigns scores to pairs of examples (Flach & Matsubara, 2007), but does not specifically consider finding the optimal pairwise ranker.)

Our approach is related to the *single-index model* (Manski, 1975) class of probabilities, $\Pr[y = 1|x] = f(w^T x)$, where $f(\cdot)$ is an *unknown* link function, in contrast to a generalized linear model which assumes a *specific* link function. The *isotonic single-index model* is where $f(\cdot)$ is assumed to be monotone increasing. Many existing methods to learn single-index models rely on some form of iteration between optimizing for $w$ and learning $f(\cdot)$. For example, the recent Isotron algorithm (Kalai & Sastry, 2009) also uses isotonic regression to provably learn single index models, and relies on alternately updating $w$ via a perceptron-like update, and running PAV to learn $f(\cdot)$. Our approach does not have similar generalization bounds, but is more direct and time-efficient, as it requires only a single call to the PAV algorithm.

## 5. Experimental results

Our experiments aim to study the conditions under which our method may improve performance over linear or logistic regression, both on synthetic and real-world datasets.

### 5.1. Methods compared

We denote our method by **Rank + IR**. For comparison, we used linear (**LinReg**) and logistic (**LogReg**) regression, as well as the results of post-processing these methods with isotonic regression. We also used the combined regression and ranking model (**CRR**) of Sculley (2010). We do not post-process CRR because that framework is explicitly designed with the aim of providing a good ranking as well as regression, which we would like to compare to our approach; our hypothesis is that our method should provide the most accurate probabilities, while additionally providing an equitable ranking to the CRR model.

Following Sculley (2010), we use the pairwise ranking framework (Herbrich et al., 2000; Joachims, 2002) with logistic loss to optimize for AUC directly, which lends itself naturally to large-scale implementation using stochastic gradient descent. For this and the CRR model, we used the Sofia-ML package[3]. All models were regularized. To test the accuracy of probability estimates, where available, we use the domain-specific metric of interest e.g. overall utility, else we measure the mean squared error between test labels and model predictions.

### 5.2. Results on synthetic dataset

We first study the performance of our proposed method on a synthetic dataset, to see the conditions under which we can expect it to improve performance over existing methods. In particular, we study the performance of various methods

---

[3] http://code.google.com/p/sofia-ml/

where the true probability model is

$$\Pr[y = 1|x; w] = a\mathbf{1}[w^T x < 0] + (1-a)\mathbf{1}[w^T x \geq 0],$$

where $0 \leq a \leq \frac{1}{2}$ controls the floor and ceiling of the probability distribution. Such capped distributions arise in e.g. item response theory (Hambleton et al., 1991), where the probability of a student answering a question correctly is bounded from below by the success rate of random guessing. Logistic regression is misspecified for this link, although for $a = 0$ the sigmoid is a reasonable approximation, while for $a = \frac{1}{2}$ the probability is independent of $x$ and thus can be modelled entirely by a bias term.

We proceed as follows: we first pick some value for $a$, and drawn $n$ samples in $\mathbb{R}^2$ from $\mathcal{N}(0, I)$. We then draw their corresponding labels, and train the various methods. We then create a separate test set through this same procedure, and evaluate the squared error of each model's predictions to the *true* probabilities of the data points (as opposed to the *labels* for these points.) We repeat the process multiple times and find the average error. We do this for $a \in \{2^{-9}, 2^{-7}, \ldots, 2^{-1}\}$.

Our results for $n = 1000$ samples are shown in Figure 1. As expected, at the endpoints of $a \to 0^+$ and $a = \frac{1}{2}$, we see that there is not much to choose between the methods. However, for intermediate values of $a$, logistic regression's performance severely deteriorates. Post-processing these scores with isotonic regression reliably estimates the floor and ceiling of the link function, and significantly improves performance. Using our method, where we post-process the scores obtained from a ranking loss, we get a small further boost in performance.

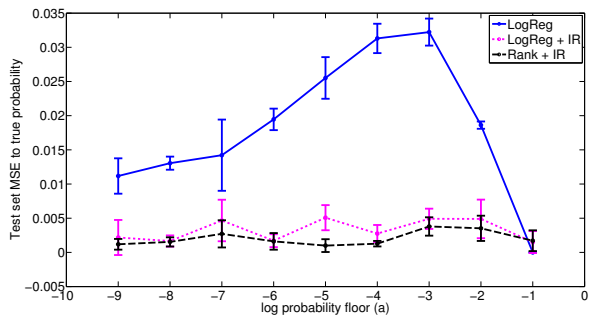

*Figure 1.* Results on synthetic dataset.

### 5.3. Results on real-world datasets

We provide experimental results on datasets drawn from the three motivating problems described in Section 2.

**Hospital Discharge**. The first dataset is from medical informatics (El-Kareh et al., 2010), where the goal is to predict follow-up errors on microbiology cultures. Predicting the probability of an example having a follow-up error helps an expert determine an appropriate action to take.

There are 8668 examples with 10 features, and we create 20 random $80 - 20$ train-test splits. Table 1 shows that our method does manage to achieve both good regression and ranking performance. Interestingly, isotonic regression slightly worsens the MSE for both linear and logistic regression, suggesting that the majority of the error arises from the basic parametric model for ranking examples itself, rather than the choice of link function.

Table 1. Average test split results on Hospital Discharge dataset.

| Method | MSE | AUC |
| --- | --- | --- |
| LinReg | $0.0461 \pm 0.0000$ | $0.6987 \pm 0.0013$ |
| LinReg + IR | $0.0465 \pm 0.0002$ | $0.6987 \pm 0.0013$ |
| LogReg | $0.0458 \pm 0.0001$ | $0.7066 \pm 0.0009$ |
| LogReg + IR | $0.0461 \pm 0.0001$ | $0.7066 \pm 0.0009$ |
| CRR | $0.0461 \pm 0.0000$ | $0.7045 \pm 0.0016$ |
| Rank + IR | $0.0460 \pm 0.0003$ | $0.7081 \pm 0.0021$ |

**KDDCup '98.** The second dataset is from the 1998 KDD Cup[4]. Here, the goal is to predict how much a individual will donate, so as to decide whether to contact them for a mail campaign (which costs money). The final utility measure is the expected profit in dollars if one contacts all individuals that the model predicts will donate (the profit takes into account the cost of contacting each individual). The data consists of $95,412$ training examples and $96,367$ test examples. We follow the strategy of (Zadrozny & Elkan, 2001): we selected the 15 features it recommends, compute the probability an individual will respond to the campaign, and then compute the expected donation given a response.

Table 2 summarizes the utility of the compared methods, as well as the AUC for the label of whether a person donates or not, on the provided test set. Our method gets an additional profit of around $300 over logistic regression, along with a small improvement in AUC. Such additional revenue may be important in practice, especially with a larger pool of candidate donors. (Note that IR sometimes modifies AUC of the input model; this is because regularization strength is picked based on utility, rather than AUC.)

Table 2. Test set results on KDDCup '98 dataset.

| Method | Test set profit | AUC |
| --- | --- | --- |
| LinReg | $12,479.12 | 0.6157 |
| LinReg + IR | $13,142.72 | 0.6157 |
| LogReg | $13,338.22 | 0.6160 |
| LogReg + IR | $12,861.88 | 0.6160 |
| CRR | $13,249.60 | 0.6162 |
| Rank + IR | $13,671.44 | 0.6162 |

**GCAT.** Lastly, we consider a classification scenario where the training set comprises only positive and unlabelled data. Based on (Elkan & Noto, 2008), one way to solve this is to predict the probability of an example being labelled, call this $\Pr[l = 1|x]$, based on which we can estimate the probability that it is positive by the identity $\Pr[y = 1|x] = \Pr[l = 1|x]/c$, where $c = \Pr[l = 1|y = 1]$ may be estimated by taking the average value of $\Pr[l = 1|x]$ on the positive examples. We simulate this scenario on the GCAT dataset[5], comprising $23,149$ examples and $47,236$ features: we construct a training set by first picking 30% of the positives (which are assigned a positive label), and then 80% of the other examples (which are treated as unlabelled). We report the primary error measures in this problem, MSE and AUC in distinguishing positive versus *negative* examples.

Table 3 summarizes the results from 20 random train-test splits. We see that post-processing logistic regression significantly improves the MSE performance over logistic regression and CRR, indicating the sigmoid link function is misspecified for this problem. Our method manages to further improve MSE, while achieving equitable ranking to other methods.

Table 3. Average test split results on GCAT dataset.

| Method | MSE | AUC |
| --- | --- | --- |
| LinReg | $0.0550 \pm 0.0015$ | $0.9824 \pm 0.0017$ |
| LinReg + IR | $0.0478 \pm 0.0021$ | $0.9823 \pm 0.0014$ |
| LogReg | $0.0579 \pm 0.0021$ | $0.9836 \pm 0.0007$ |
| LogReg + IR | $0.0423 \pm 0.0024$ | $0.9836 \pm 0.0007$ |
| CRR | $0.0557 \pm 0.0020$ | $0.9825 \pm 0.0015$ |
| Rank + IR | $0.0419 \pm 0.0021$ | $0.9831 \pm 0.0005$ |

Overall, on all three datasets, we see our method achieves both good ranking and regression performance, and on the KDDCup and GCAT datasets manages to improve overall regression performance. Note that logistic and linear regression are strong baselines, and that even small improvements in performance may be significant in practical applications (Sculley, 2010).

## 6. Conclusion and future work

Many real-world applications of predictive models require predicting accurate probabilities of class membership. We studied the principles behind predicting accurate probabilities, and proposed a simple method to achieve it. Our method is based on post-processing the results of a model that optimizes a *ranking* loss with isotonic regression. The model is shown to have good empirical performance. In the future, it would be interesting to study the theoretical properties of the model more closely, and evaluate the model in other scenarios requiring probability estimates.

## Acknowledgements

XJ and LOM were funded in part by the National Library of Medicine (R01LM009520) and NHLBI (U54 HL10846).

---

[4] http://www.kdnuggets.com/meetings/kdd98/kdd-cup-98.html

[5] http://vikas.sindhwani.org/svmlin.html